\documentclass[sigconf,9pt]{acmart}

\AtBeginDocument{%
  }

\setcopyright{acmlicensed}
\copyrightyear{2018}
\acmYear{2018}
\acmDOI{XXXXXXX.XXXXXXX}

\acmConference[Conference acronym 'XX]{Make sure to enter the correct
  conference title from your rights confirmation emai}{June 03--05,
  2018}{Woodstock, NY}
\acmISBN{978-1-4503-XXXX-X/18/06}

\usepackage{tabularx}
\usepackage{lipsum}
\usepackage{graphicx}
\usepackage{subfigure}
\usepackage{enumitem}
\usepackage{wrapfig}
\usepackage{pifont}
\usepackage[ruled,vlined]{algorithm2e}
\usepackage{mathtools}

\usepackage{amsmath}
\usepackage{soul}
\usepackage{multirow}

\newcommand{\bo}[1]{\textcolor{blue}{(Bo):{#1}}}

\newcommand{\ms}[1]{\textcolor{olive}{(Manavjeet):{#1}}}

\newcommand{\test}[1]{\textit{placeholder}}
\usepackage[ruled,vlined]{algorithm2e}

\usepackage[capitalize]{cleveref}
\crefname{section}{Sec.}{Secs.}
\Crefname{section}{Section}{Sections}
\Crefname{table}{Table}{Tables}
\crefname{table}{Tab.}{Tabs.}

\newcommand{\abrtwo}[1]{\textit{EdgeSimShare}}

\begin{document}



\title{Representation Similarity: A Better Guidance of DNN Layer Sharing for Edge Computing without Training}

\author{Bryan Bo Cao, \hspace{1pt} Abhinav Sharma, \hspace{1pt} Manavjeet Singh, \hspace{1pt} Anshul Gandhi, \hspace{1pt} Samir Das, \hspace{1pt} Shubham Jain}

\affiliation{\institution{Stony Brook University}
\country{}
}

\affiliation{\{boccao,abhinsharma,manavsingh,anshul,samir,jain\}@cs.stonybrook.edu
\country{}
}


\begin{abstract}
Edge computing has emerged as an alternative to reduce transmission and processing delay and preserve privacy of the video streams.
However, the ever-increasing complexity of Deep Neural Networks (DNNs) used in video-based applications (e.g. object detection) exerts pressure on memory-constrained edge devices. 
\textit{Model merging} is proposed to reduce the DNNs' memory footprint by keeping only one copy of merged layers' weights in memory.
In existing model merging techniques, (i) only architecturally identical layers 
can be shared;
(ii) requires computationally expensive retraining in the cloud; 
(iii) assumes the availability of ground truth for retraining.
The re-evaluation of a merged model's performance, however, requires a validation dataset with ground truth, typically runs at the cloud. Common metrics to guide the selection of shared layers include the size or computational cost of shared layers or representation size. We propose a new model merging scheme by sharing representations (i.e., outputs of layers) at the edge, guided by representation similarity $S$. We show that $S$ is extremely highly correlated with merged model's accuracy with Pearson Correlation Coefficient $|r| > 0.94$ than other metrics, demonstrating that representation similarity can serve as a strong validation accuracy indicator without ground truth. We present our preliminary results of the newly proposed model merging scheme with identified challenges, demonstrating a promising research future direction. 


\end{abstract}

\acmYear{2024}\copyrightyear{2024}
\setcopyright{acmlicensed}
\acmConference[ACM MobiCom '24]{The 30th Annual International Conference on Mobile Computing and Networking}{November 18--22, 2024}{Washington D.C., DC, USA}
\acmBooktitle{The 30th Annual International Conference on Mobile Computing and Networking (ACM MobiCom '24), November 18--22, 2024, Washington D.C., DC, USA}
\acmDOI{10.1145/3636534.3695903}
\acmISBN{979-8-4007-0489-5/24/11}

\maketitle


\section{Introduction}



Globally, the video analytics market is growing at a swift pace. It is expected to grow to a net worth of USD 22.6 billion in 2028, from USD 8.3 in 2023 \cite{cagr}. 
Traditionally, video analytic frameworks stream camera feeds to the cloud data centers for processing.
While it provides access to sufficiently powerful processors, streaming to cloud data centers comes with increased latency, subscription fees, potential privacy concerns, and in some regions, government policies prohibiting it entirely.
Edge computing offers real-time processing solutions by using computing resources located closer to the data source enhancing response times.
The edge local approach is crucial for applications requiring swift action, such as urban traffic management, security system monitoring, and autonomous vehicle navigation 
.

\begin{figure}[t]
    \centering
    \includegraphics[width=0.85\linewidth]{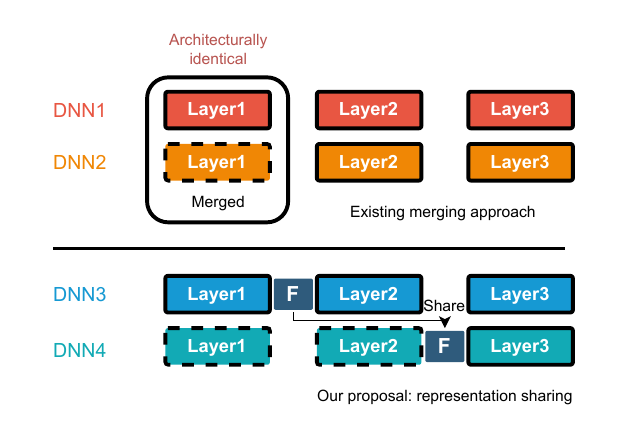}
    \vspace{-18pt}
    \caption{Proposed merging scheme: sharing representations. DNNs are denoted in different colors. Layers with solid borders represent the weights are loaded in memory while dashed borders indicate the layer weights are offloaded. F: representation.} 
    \Description{Proposed merging scheme: sharing representation. DNNs are denoted in different colors. Layers with solid borders represent the weights are loaded in memory while dashed borders indicate the layer weights are offloaded. F: representation.} 
    \label{fig:summary}
\end{figure}

Edge computing relies on embedded devices, including but not limited to GPU-accelerated developer kits such as Jetson Nanos and Jetson Orins. 
Although edge computing is a lucrative alternative to the cloud, it is extremely restricted in GPU memory, 
making it impossible to load multiple or sometimes even a single model. For example, due to its limited memory, a TensorRT implementation of yolov8n-pose \cite{yolov8_ultralytics} cannot be loaded on Jetson Nano. 
A technique to overcome memory limitations is model merging.
Model merging combines architecturally identical layers from different Deep Neural Networks (DNNs), and has been proposed to reduce memory usage and enable concurrent operation on limited-memory edge devices while maintaining performance.
Prior work on merging technique \cite{padmanabhan2023gemel} considers only architectural similarity, i.e., the candidate layers to be merged have to follow strictly the same definitions.
Merging DNNs during inference at the edge without retraining has been far less explored for video analytics. 
To the best of our knowledge, we are the first to explore the edge GPU memory restriction problem under the following constraints: (1) \textit{cloud servers} and (2) \textit{ground truth} are unavailable.

\begin{figure*}[t]
  \begin{minipage}{1\linewidth}
  \centering
    \subfigure[Same-Stage Stage3]{\includegraphics[width=0.3\textwidth]{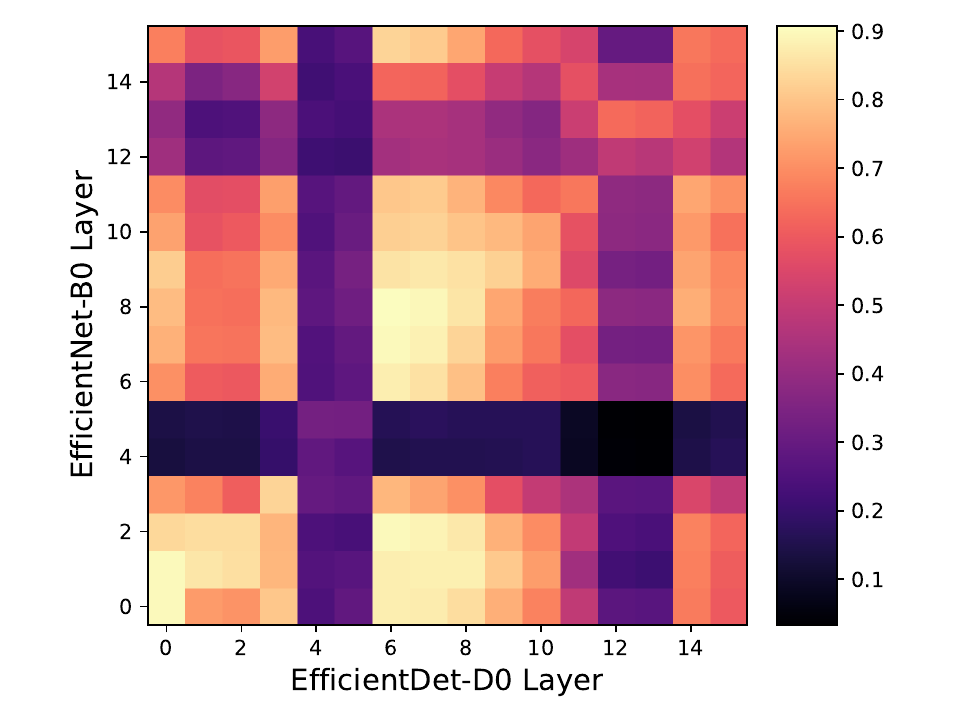}}
    \subfigure[Same-Stage Stage7]{\includegraphics[width=0.3\textwidth]{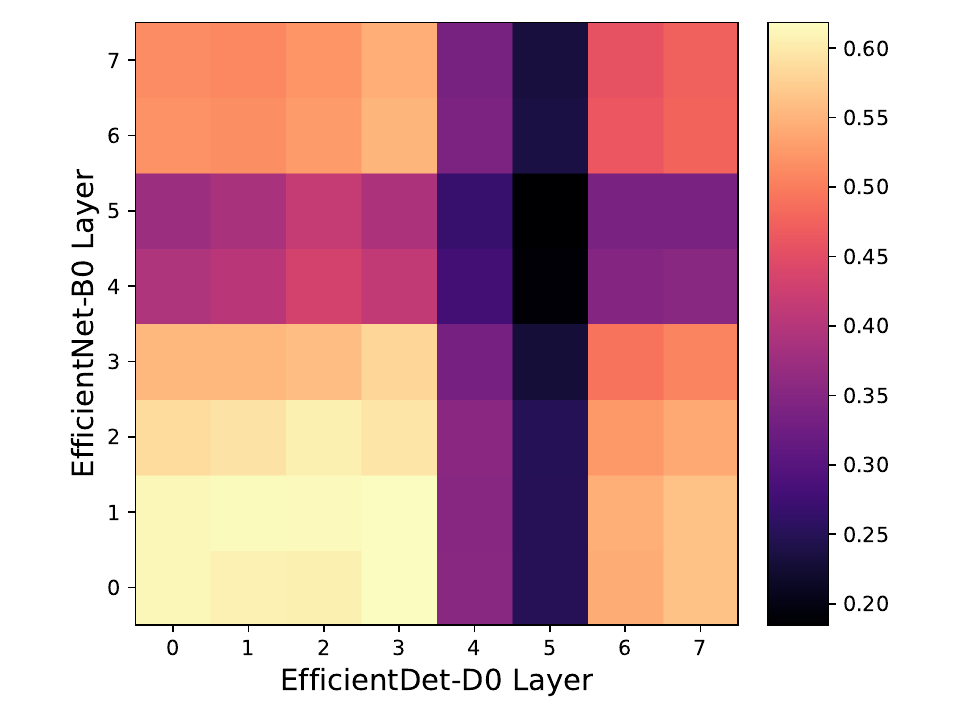}}
    \subfigure[Cross-Stage]{\includegraphics[width=0.3\textwidth]{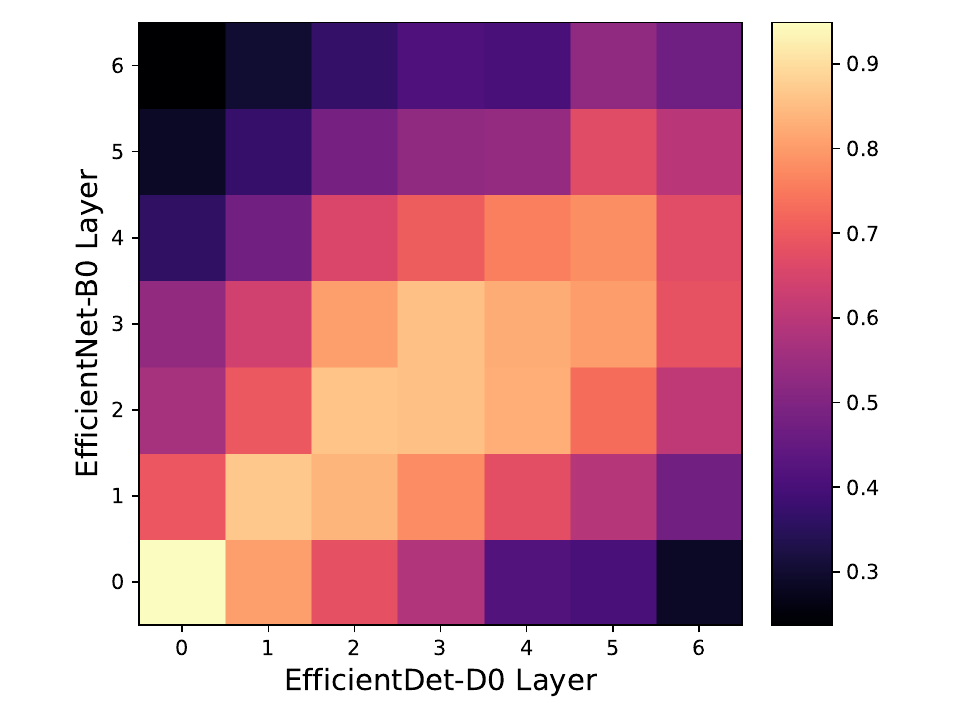}}
  \vspace{-9pt}
  \caption{Same-Stage (a) (b) and Cross-Stage (c) representation similarity heatmaps for EfficientNet-B0 and EfficientDet-D0. Stages with higher similarity are visualized in brighter colors while darker ones depict dissimilar representations.}
  \Description{Same-Stage (a) (b) and Cross-Stage (c) representation similarity heat maps for EfficientNet-B0 and EfficientDet-D0. Stages with higher similarity are visualized in brighter colors while darker ones depict dissimilar representations.}
  
  \label{fig:sim_acc}
  \end{minipage}
\end{figure*}

We propose sharing representations for model merging and leverage representation similarity to estimate merged model's accuracy. Our findings in the pivot study demonstrate that DNNs can preserve performance as long as the received feature maps (representations) of a layer are close, or similar enough to its original ones without merging, shown in Fig. \ref{fig:summary}.

For the first time, this breaks the architectural similarity assumption in previous works \cite{padmanabhan2023gemel, jiang2018mainstream} and opens a new layer-merging design space where merging can be applied at any layer -- as long as the representations are similar between two merged candidate layers.
In this paper, we identify the limitations of existing methods that impede a more resource-efficient merging approach, specifically with the assumptions of (1) architecturally identical layers (2) training at the cloud, and (3) the need for ground truth to compute merged model accuracy. We propose a novel approach of model merging to break these assumptions by leveraging representation similarity computed by Centered Kernel Alignment (CKA) \cite{nguyen2020wide}. To the best of our knowledge, we are the first to explore sharing DNN layers without requiring architecturally identical layers, avoiding cloud-based training, and without ground truth for post-sharing performance estimation. We have met the following challenges:
\begin{enumerate}
    \item \textbf{Non-Sequential DNN layers.} Unlike sequential layers in a single path of input and output, non-sequential layers (such as skip-connections) have to ensure the completeness of received inputs from all previous paths, requiring a mechanism to store the representations from other paths.
 
    \item \textbf{Mismatched representation shapes for sharing.} When merging layers, the shapes of the input and output for the merged layer have to be matched with the previous ones. Without architecturally identical layers, these representation shapes can be different.


    \item \textbf{Lack of guidance for layer sharing.} Selecting layers for merging depends on some metrics such as computation cost, or layer size to minimize memory footprint. However, the extent of accuracy loss due to merging is unknown, resulting in ineffective guidance for layer sharing.
    
    
\end{enumerate}





\section{Approach}
In DNNs, model architecture defines the structural layout and connections between different types of layers, determining the flow of data through the model. Weights are the adjustable parameters within this architecture, updated during training to minimize loss and influence the model's outputs based on input features.

Edge devices benefit from model merging \cite{padmanabhan2023gemel} since only one copy of the weights of layers from two or more models is hosted in memory. Instead of merging architecturally identical layers in prior works \cite{padmanabhan2023gemel,jiang2018mainstream}, we propose \textbf{sharing representations} of merged layers. We assume the weights of the layers before the shared representations in the target model are not loaded into memory. Sharing representations allows more flexible layer merging schemes, such as in the earlier layers from one model to deeper layers in another model. In this work, we study stage-wise representation sharing. A stage consists of several convolutional layers of the same type.


\noindent \textbf{Measuring Layer Similarity Using CKA.} Our key idea centers on the hypothesis that merged model's accuracy is proportional to layer similarity -- the similarity of merged candidate layers from two models before merging. We utilize representation similarity \cite{nguyen2020wide} by focusing on the outputs of intermediate layers as a means of quantifying layer similarity.

Formally, we denote two layers' representations as $X$ and $Y$. The similarity of two representations is computed by Equation \ref{equ:cka}:
\vspace{-1pt}
\begin{equation}
\text{CKA}(K, L) = \frac{\text{HSIC}_0(K, L)}{\sqrt{\text{HSIC}_0(K, K) \cdot \text{HSIC}_0(L, L)}}
\label{equ:cka}
\end{equation}
\vspace{-1pt}where the similarity of a pair of examples in $X$ or $Y$ is encoded in one element of the Gram matrices $K = XX^{T}$ and $L = YY^{T}$. The similarity of these centered similarity matrices is captured by $HSIC$, invariant to representations' orthogonal transformations. We leverage this key property to measure mismatched shape representations for cross-layer sharing. In the following sections, representation similarity is referred to as $S$. Fig. \ref{fig:sim_acc} shows stage-wise representation similarity for Same-Stage (a), sharing representations from the same stage and Cross-Stage (b) in cross stages.

\noindent \textbf{Representation Shape Consistency for Merged Models.}
To tackle the challenge of non-sequential DNN layers (such as skip-connections in EfficientDet-D0) we insert \emph{None} tensors at the start of inputs corresponding to each layer of later segments of the original model, concatenated with the representations from the previous layer. These \emph{None} tensors shift the indices of incoming data to align with their expected positions in an unsharded model, thereby maintaining the structural and functional integrity of the model post-sharding. 
In the scenario of mismatched representation shapes, we note two representations' shapes as $(C_{i}^{a}, H_{i}^{a}, W_{i}^{a})_{s}$ and $(C_{j}^{b}, H_{j}^{b}, W_{j}^{b})_{t}$ where $C$ is the number of channels, $H$ and $W$ are the resolution dimensions of a representation, $i$ and $j$ denote the layer indices, $a$ and $b$ represent two models, $s$ represents the shared representation and $t$ is the target representation.
To match $s$ with $t$ for sharing, when $C_{i}^{a} \neq C_{j}^{b}$, we uniformly sample $C_{i}^{b}$ representations from $C_{i}^{a}$ with repetition. 
To match resolutions, we upsample $(H_{i}^{a}, W_{i}^{a})_{s}$ to $(H_{j}^{b}, W_{j}^{b})_{t}$ if the former is smaller than the latter; otherwise downsample $(H_{i}^{a}, W_{i}^{a})_{s}$ to $(H_{j}^{b}, W_{j}^{b})_{t}$.

\section{Evaluation}

We use EfficientNet-B0 and EfficientDet-D0 pre-trained models in our evaluations, due to the common use case of image classification (IC) and object detection (OD). 
A naive way to run multiple application pipelines (e.g., IC and OD) is to load both models into GPU memory. When representations are shared, only one copy of the weights from that layer is loaded instead of two.

We conduct stage-wise experiments and measure the metrics in each stage, including validation accuracy (Acc.), the conventional metrics of floating point operations (FLOPs), representation size (Size), and number of parameters ($\#$Params), as well as our proposed representation similarity (S). In each experiment, we share the representations in one stage (the outputs of the last layer) from EfficientDet-D0 with EfficientNet-B0. Representation similarity is computed using the same inputs from the validation set. We denote a merged model's accuracy as the validation accuracy on ImageNet for classification after sharing the representations of one stage in EfficientDet-D0 with EfficientNet-B0.
Next, we quantify the correlation between the merged model's classification accuracy and the rest of metrics by Pearson Correlation Coefficient \cite{schober2018correlation}, denoted as $r$. The objective is to investigate to what extent representation similarity can be used to estimate a merged model's accuracy after sharing, which can provide better guidance in future work.

Our experiments are categorized in two types: (1) Same-Stage Layer Sharing and (2) Cross-Stage Layer Sharing.

\begin{figure}[ht]
    \centering
    \includegraphics[width=0.99\linewidth]{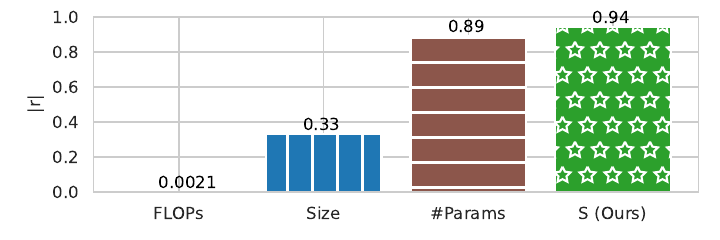}
    \vspace{-9pt}
    \caption{Stage-Wise absolute value of Pearson Correlation Coefficient $r$ between merged accuracy (Acc.) and representation similarity (S) using different metrics. Results indicate a highly correlated relationship ($|r|$ = 0.94) between Acc. and $S$.}
    \label{fig:allcorr}
\end{figure}

\begin{figure}[h]
  \centering
    \subfigure[Same-Stage]{\includegraphics[width=0.405\linewidth]{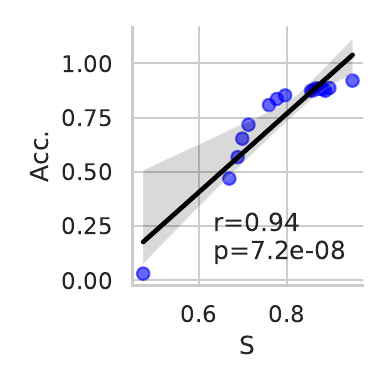}}
    \hspace{8pt}
    \subfigure[Cross-Stage]{\includegraphics[width=0.405\linewidth]{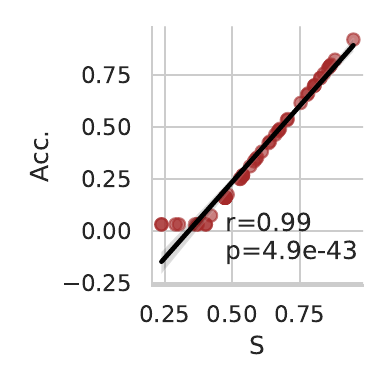}}
  \vspace{-9pt}
  \caption{Pearson Correlation Coefficient $r$ between accuracy (Acc.) and representation similarity (S). Each dot denotes a merged model's Acc. and S after representation sharing. (a): Same-Stage similarity; (b): Cross-Stage similarity.}
  \Description{Pearson Correlation Coefficient $r$ between accuracy (Acc.) and representation similarity (S). Each dot denotes a merged model's Acc. and S after representation sharing. (a): Same Layer similarity; (b): Cross Layer similarity.}
  \label{fig:simcorr}
\end{figure}

\noindent \textbf{Same-Stage Layer Sharing.} In each experiment, we share the representations of one stage in EfficientDet-D0 with EfficientNet-B0 in the same stage. We use the absolute value of $|r|$ in this experiment, as the magnitude denotes the degree of correlation. In Fig. \ref{fig:allcorr} we demonstrate that merged model accuracy (Acc.) is highly correlated with representation similarity ($S$) $|r|=0.94$ (detailed in Fig. \ref{fig:simcorr} (a)), compared to $\#$Params ($|r|=0.89$), Size ($|r|=0.33|$) and FLOPs ($|r|=0.0021$). An interesting observation is that there exists a threshold $S^{\prime}=0.4$ where the accuracy is extremely low as $0.031$ when $S < S^{\prime}$ while the relationship between Acc. and S is almost linear when $S > S^{\prime}$. This further confirms the feasibility of Cross-Layer sharing that allows for more flexible layer sharing schemes.

\noindent \textbf{Cross-Stage Layer Sharing.} We enumerate all pairs of stages between EfficientDet-D0 and EfficientNet-B0, and share the representations of one stage in EfficientDet-D0 with EfficientNet-B0 in a different stage. Results in Fig. \ref{fig:simcorr} (b) depict a strong correlation between merged model accuracy (Acc.) and representation similarity (S) across stages with $|r|=0.99$. For the first time, we demonstrate the feasibility of breaking the architectural identity assumption.

\section{Conclusion and Future Work}
We have explored and presented the first study of DNN layer sharing guided by representation similarity. Our results show a strong correlation between representation similarity and merged model accuracy with Pearson Correlation Coefficient $|r| \geq 0.94$, demonstrating representation similarity as an accurate indicator for merged model accuracy estimation.
In addition, we push the boundary from same-layer sharing to cross-layer sharing.

We are designing and implementing the rest of the system from the following perspectives, aiming for real-world deployment on edge devices: (1) multiple video pipelines for different tasks, such as pose estimation, license plate recognition, and so forth; (2) algorithms searching for the optimal sharing point jointly considering resource consumption and representation similarity; (3) methods for preserving post-sharing accuracy without ground truth.


\section{Acknowledgements}
This research has been supported in part by the National Science Foundation (NSF) under Grant No. CNS-2055520 and CNS-2214980.

\bibliographystyle{ACM-Reference-Format}
\bibliography{main.bib}

\appendix

\end{document}